\documentclass[final]{l4dc2025}

\title[Reinforcement Learning from Multi-level and Episodic Human
Feedback]{Reinforcement Learning from Multi-level and Episodic \\Human
Feedback}
\usepackage{times}

\coltauthor{\Name{Muhammad Qasim Elahi\textsuperscript{*}} \Email{elahi0@purdue.edu}\\
\Name{Somtochukwu Oguchienti\textsuperscript{*}} \Email{soguchie@purdue.edu}\\
\Name{Maheed H. Ahmed} \Email{ahmed237@purdue.edu}\\
\Name{Mahsa Ghasemi} \Email{mahsa@purdue.edu}\\
\addr Elmore Family School of Electrical and Computer Engineering, Purdue University, West Lafayette, IN 47907}

\usepackage{bbm}
\usepackage{amsmath}
\usepackage{amssymb}
\usepackage[ ]{algorithmic}

\usepackage{graphicx}
\usepackage{caption}
\usepackage{subcaption}
\usepackage[symbol]{footmisc}

\DeclareMathOperator*{\argminB}{argmin}   

\newtheorem{assumption}{Assumption}

\begin{document}

\maketitle

\begin{abstract}%
Designing an effective reward function has long been a challenge in reinforcement learning, particularly for complex tasks in unstructured environments. To address this, various learning paradigms have emerged that leverage different forms of human input to specify or refine the reward function. Reinforcement learning from human feedback is a prominent approach that utilizes human comparative feedback—expressed as a preference for one behavior over another—to tackle this problem. In contrast to comparative feedback, we explore multi-level human feedback, which is provided in the form of a score at the end of each episode. This type of feedback offers more coarse but informative signals about the underlying reward function than binary feedback. Additionally, it can handle non-Markovian rewards, as it is based on an entire episode's evaluation. We propose an algorithm to efficiently learn both the reward function and the optimal policy from this form of feedback. Moreover, we show that the proposed algorithm achieves sublinear regret and demonstrate its empirical effectiveness through extensive simulations.
\end{abstract}


\section{INTRODUCTION}
Reinforcement Learning (RL), a framework for addressing the problem of decision-making under uncertainty, has proven to be highly effective in tackling complex real-world challenges. In the traditional RL setting, the agent interacts with the environment and receives a reward signal for each state-action pair. This paradigm has achieved huge successes in real-life applications ranging from games \cite{mnih2013playing, mnih2015human, silver2016mastering, silver2018general}, manufacturing \cite{wang2005application, liu2022deep}, medicine \cite{zhao2011reinforcement, coronato2020reinforcement}.
\footnote[0]{*Equal Contribution}

This conventional setting assumes that a reward signal is readily available to the agent and accurately specifies the objective of the task. However, many real-world settings lack a clear reward signal for each state-action pair. For instance, in tasks like autonomous driving or stacking books on a shelf, defining a precise reward is complex, but it is feasible to evaluate success at the end of the task. Therefore, engineering a reward is very difficult and poses a significant challenge to RL. To address this challenge, \cite{eysenbach2018diversity} proposes methods that do not rely on predefined reward functions to learn diverse skills. Some earlier works explore Inverse Reinforcement Learning (IRL) as an approach to learn a reward function from demonstrations of successful trajectories \cite{abbeel2004apprenticeship, ziebart2008maximum}. However, IRL depend heavily on high-quality demonstrations, which can be costly and difficult to obtain. Recently, Reinforcement Learning from Human Feedback (RLHF) has emerged as a promising approach, particularly in human-robot interaction settings, where human feedback replaces the need for explicit reward functions \cite{knox2008tamer, macglashan2017interactive}.
Within the framework of RLHF, \cite{christiano2017deep} proposes pairwise comparison of trajectories while \cite{rendle2012bpr,brown2019extrapolating} leverage ranking of set of trajectories. Additionally, several techniques strategically combine both demonstrations and preferences to learn the reward function \cite{biyik2022learning}. However, feedback from preferences is typically less informative than scalar reward functions, which poses a challenge for RLHF. Moreover, humans can be biased in their preferences, hence preferences may not capture the true reward signal.

Addressing the limitations of IRL and RLHF, \cite{agarwal2019learning} proposes RL from sparse rewards where feedback is only provided at the end of a trajectory. This technique has shown potential in achieving optimal policies with low regret. However, the authors consider a setting where only binary human feedback scores are available \cite{chatterji2021theory}.
In contrast, we examine a richer feedback structure, where the agent receives different levels of feedback, ranging from $0$ to $K-1$, instead of a binary feedback. We propose an algorithm that leverages such categorical feedback to learn the underlying reward function. The main contributions of our work are as follows:

\begin{itemize}
    \item We propose an online optimism-based algorithm that leverages multi-level and episodic feedback signals to estimate the underlying reward function and, consequently, learn the optimal policy for an episodic MDP setting.
    \item We analyze the proposed algorithm and show that it achieves sublinear regret in terms of the number of episodes under mild regularity assumptions.
    \item We show experimentally, using series of grid-world simulations, that our algorithm is able to learn an optimal policy from multilevel feedback. 
\end{itemize}

\section{RELATED WORK}
Preference-based reinforcement learning (PbRL) has been extensively studied empirically, with diverse applications in domains such as games \cite{ibarz2018reward, wirth2014learning, runarsson2014preference}, robotics \cite{jain2013learning, kupcsik2018learning} and training of large language models \cite{stiennon2020learning, ouyang2022training}. These empirical successes have motivated theoretical analysis of PbRL, providing provable guarantees for learning near-optimal policies across different settings. \cite{pacchiano2021dueling, zhan2023provable, novoseller2020dueling} explore the online setting under linear or tabular MDPs for learning near-optimal policies from human preferences. In the offline setting, \cite{shin2023benchmarks, zhan2023provableb, zhu2023principled} show that near-optimal policies can be learned from collected human preference data under the maximum likelihood framework. Extending beyond tabular MDPs, \cite{chen2022human} explores PbRL with general function approximation. While \cite{novoseller2020dueling, pacchiano2021dueling} assume an underlying utility function generating human preferences, \cite{dudik2015contextual, wang2023rlhf} adopt more general preference models based on the von Neumann winner policy. More recent studies explore preference feedback via pairwise comparisons \cite{novoseller2020dueling, wirth2016model} or a score-based feedback \cite{efroni2021reinforcement, chatterji2021theory}.

IRL techniques assume the existence of expert demonstrations to learn a reward function for the agent \cite{abbeel2004apprenticeship}. Hence, the policy derived from the learned reward function is inherently limited by the quality of the collected demonstrations. In practice, expert demonstrations are difficult to obtain due to the complexity of the environment or the demonstrator's inexperience, which can limit the quality of their demonstrations. This challenge motivated recent studies to address the problem of suboptimality in the human demonstrations \cite{brown2019extrapolating, oguchienti2023inverse, chen2021learning}. \cite{brown2019extrapolating} proposes T-REX algorithm to learn an optimal policy from trajectory rankings. Building on this, \cite{brown2020better, chen2021learning} design a learning framework using noise-injected policies to learn a reward function. Assessing demonstrator expertise, \cite{beliaev2022imitation} explores an imitation learning setting to learn an optimal policy alongside the expertise level corresponding to each demonstrator. Similarly, \cite{oguchienti2023inverse} extends to an IRL setting to learn the expertise levels of each demonstrator and incorporates this parameter into the reward learning framework. Furthermore, \cite{beliaev2024inverse} models suboptimality in human demonstrations via their reward bias and action variance and proposes an algorithm to learn an optimal policy alongside these parameters. 

Online RL focuses on techniques that achieve low regret for agent's interaction while balancing exploration and  exploitation based on the principle of optimism in the face of uncertainty. Several authors have analyzed the sample complexity of online RL in episodic MDPs. For instance, \cite{osband2016lower, azar2017minimax, pacchiano2021towards} provide theoretical guarantees for regret bounds in this setting. Additionally, \cite{auer2006logarithmic, auer2008near} propose the upper confidence reinforcement learning (UCRL) algorithm for undiscounted MDPs, aimed at deriving regret bounds over a finite number of steps. \cite{liu2020regret, zhou2021provably} provide similar analysis for discounted MDP settings. Moreover, \cite{azar2017minimax} develop the upper confidence bound value iteration (UCB-VI) algorithm to address the problem of optimal exploration in finite-horizon MDPs.  Beyond UCB-based approaches, \cite{osband2013more, agrawal2017optimistic} explore posterior sampling based RL (PSRL) algorithm that samples from the posterior distribution over MDPs and chooses the optimal actions based on the sampled MDP. Similar to \cite{chatterji2021theory}, our work adopts the UCB-based framework for non-Markovian rewards. However, \cite{chatterji2021theory} assumes that the reward is drawn from an unknown binary logistic model. This model is inherently limited since the agent receives a score of $0$ or $1$ indicating whether a goal is achieved or not. To address this limitation, we propose a more
informative feedback model based on a categorical distribution, specifically the softmax distribution. This approach allows the agent to receive a richer feedback based on the observed trajectory,
overcoming the binary feedback constraint.

\section{PRELIMINARIES}

A Markov Decision Process (MDP) in an episodic setting is represented as \( \mathcal{M} = (\mathcal{S}, \mathcal{A}, \mathbb{P}, H, \rho) \), where \( \mathcal{S} \) is a finite set of states, \( \mathcal{A} \) is a finite set of actions, \( \mathbb{P}(\cdot | s, a) \) represents the state transition distribution, and \( H \) is the horizon or length of trajectories. The initial state \( s_0 \) is sampled from a known distribution \( \rho \). We assume that all state transition probabilities under all actions are known. A trajectory \( \tau  = (s_0,a_0, ..., a_{H-1},s_H)\)  consists of all states and actions for a given episode, and we denote the set of all possible trajectories by \( \mathcal{T} \). 
Our formulation differs from the standard RL setting, where the agent receives a reward for every state-action pair, i.e., \( R(s, a) \). Instead, the reward is a function of the entire trajectory, \( R(\tau) \). A reward that is a function of the entire trajectory can capture non-Markovian tasks by incorporating historical context. We propose \(K\)-ary trajectory feedback labels, where the agent receives a categorical score rather than a binary score. \(K\)-ary feedback is easier for humans to provide, serves as an abstraction of real-valued rewards, and is more expressive than binary feedback as used in \cite{chatterji2021theory}.  Let \( \mathbf{w}_i^\star \in \mathbb{R}^d, \; \forall i \in \{0, 1, \dots, K-1\} \) be a set of \( K \) vectors unknown to the learner. Furthermore, every trajectory \( \tau \) is associated with a known feature vector \( \phi(\tau) \in \mathbb{R}^d \).

\begin{assumption} [K-ary Categorical Feedback] For any trajectory $\tau \in \mathcal{T}$, the agent receives  categorical feedback $y_{\tau} \in \{0,1,2,...K-1 \}$ with the probability distribution as follows:
\begin{equation*}
        P(y_\tau = i)= \frac{\exp(\mathbf{w}_{i}^{\star T}\phi(\tau))} {\sum^{K-1}_{j=0}\exp(\mathbf{w}_j^{\star T}\phi(\tau))},
\end{equation*}
where $i  \in \{0,1,2,...K-1 \}$.
\label{feedback_assum}
\end{assumption}

\begin{assumption}[Bounded Parameters] \label{assumption::bounded_features}
We assume that: 
\begin{itemize}
    \item $ \lVert \mathbf{w}_i^{\star} \rVert_{2} \le \frac{B}{K} \;\; \forall i \in 0,\cdots, K-1$ and for some known  $B>0$,
    \item $\|\phi(\tau)\|_2 \leq 1 \;\; \forall \tau \in \mathcal{T}$.
\end{itemize}
\end{assumption}
We make boundedness assumptions on the features and true weight parameter, which are commonly used in prior literature \cite{chatterji2021theory, zhu2023principled, faury2020improved, russac2021self}. Notice that this assumption holds without loss of generality for any finite features and finite true parameter with the appropriate choice of $B$. We denote the trajectories in the $N$ episodes by $\{\tau^{(n)}\}_{n=1}^{N}$ and the corresponding categorical feedback received by $\{y^{(n)}\}_{n=1}^{N}$. Given the stochastic nature of the reward feedback, i.e., the categorical distribution above, we aim to maximize the expected value of this reward feedback i.e. R($\tau$) = $\mathbb{E}[y_\tau]$. This is more formally expressed in the equation below:

\begin{equation}
    R(\tau) = \sum_{i=0}^{K-1} iP(y_\tau =i) =  \sum_{i=0}^{K-1} i 
 \frac{\text{exp}(\mathbf{w}_i^{\star T}\phi(\tau))} {\sum^{K-1}_{j=0}\text{exp}(\mathbf{w}_j^{\star T}\phi(\tau))}.
 \label{reward_eq}
\end{equation}
The value function of a policy $V^{\pi}(s)$ for any state $s \in \mathcal{S}$ is defined as:

\begin{equation*}
    V^{\pi}(s) := \mathbb{E}_{\substack{s_{t} \sim \mathbb{P}(.|s_{t-1},a_{t-1}) \\ a_t \sim \pi(.|s_t)}} \big[R(\tau) \;|\; S_0 = s\big] =  \mathbb{E}_{\tau \sim \mathbb{P}^{\pi}(.|s)} [R(\tau)],  
\end{equation*}

The average value function for some initial state distribution \( \rho \) is defined as \( V^{\pi} := \mathbb{E}_{s_{0} \sim \rho} [V^{\pi}(s_0)] \). The optimal policy \( \pi^{\star} \) is given by \( \pi^{\star} \in \arg \max_{\pi \in \Pi} V^{\pi} \), where \( \Pi \) is the set of all possible policies, including non-Markovian policies as well. It is worth noting that, similar to \cite{chatterji2021theory}, in our setting, the optimal policy may be \emph{non-Markovian} because of the non-Markovian reward. We use the notation \( \pi^{(n)} \) to represent the policy used for the episode \( n \). Our goal is to minimize the cumulative regret \( \mathcal{CR}(N) \) of the agent over a fixed number of episodes \( N \), which represents the cumulative suboptimality of the agent over the span of its learning/interaction.

\begin{equation}\label{def:regret}
\mathcal{CR}(N) := \sum_{n=1}^N V^{\pi^{\star}}-V^{\pi^{(n)}}.
\end{equation}

\section{LEARNING FROM MULTILEVEL FEEDBACK SCORES} \label{sec:4}

In this section, we discuss the design of our proposed algorithm that leverages multilevel episodic feedback scores and present theoretical guarantees demonstrating that the cumulative regret for our algorithm exhibits sublinear scaling with respect to the number of episodes \( N \). The details of proofs for lemmas \ref{lemm_concen}, \ref{lem2} and theorem \ref{thm1} can be found in the supplementary section of the full paper \href{https://arxiv.org/abs/2504.14732}{here}.

\subsection{Algorithm Design and Analysis}
In this subsection, we present the details of the algorithm design and provide our main theoretical results, including the cumulative regret guarantee. We propose Algorithm \ref{alg_1}, which is an adaptation of the UCBVI algorithm \cite{azar2017minimax} to our setting with multilevel episodic feedback and known transition probabilities. The learner does not have access to the true weight vectors \( \mathbf{w}_i^{\star},\; \forall i \in \{0,1,\dots,K-1\} \). Therefore, an estimate for these weight vectors must be constructed. Instead of estimating the weights individually, these vectors are concatenated into a single vector \( \mathbf{w}^{\star} \in \mathbb{R}^{Kd} \). The feature vectors for the trajectories \( (\phi(\tau)) \) are transformed into a set of vectors \( \phi_i({\tau}) \in \mathbb{R}^{Kd},\; \forall i \in \{0,1,\dots,K-1\} \). All entries of the vector \( \phi_i(\tau) \) are zeros except for those from index \( i \times d \) to \( i \times d + (d-1) \), which correspond to the entries of the feature vector \( \phi(\tau) \). Based on Assumption \ref{assumption::bounded_features}, it follows that \( ||\mathbf{w}^\star||_2 \leq B \) and \( ||\phi_i({\tau})||_2 \leq 1,\; \forall i \in \{0,1,\dots,K-1\} \). The reward for a trajectory \( R(\tau) \) can equivalently be expressed as follows:

\begin{equation*}
    R(\tau) = \sum_{i=0}^{K-1} iP(y_\tau =i) =  \sum_{i=0}^{K-1} i 
 \frac{\text{exp}(\mathbf{w}^{\star T}{}\phi_i(\tau))} {\sum^{K-1}_{j=0}\text{exp}(\mathbf{w}^{\star T}\phi_j(\tau))}.
\end{equation*}

We define feasible set for the estimated weight vector as $\mathbf{W}_B:=\{\mathbf{w} \in \mathbb{R}^{kd}\;\big{|}\; ||\mathbf{w}||_2 \leq B\}$. For any episode, say \( n \in [N] \), we construct a maximum likelihood estimate (MLE) \( \widehat{\mathbf{w}}_n \) based on the sampled trajectories \( \{\tau^{(i)}\}_{i=1}^{n} \) and the corresponding categorical feedback received as \( \{y^{(i)}\}_{i=1}^{n} \):

\begin{equation*}
    \widehat{\mathbf{w}}_n = \argminB_{\mathbf{w} \in \mathbf{W}_B} \ell^{(n)}(\mathbf{w}),
\end{equation*}

where the loss function $\ell^{(n)}(\mathbf{w})$ is defined as the negative log likelihood of the data samples, i.e., $\ell^{(n)}(\mathbf{w}) =  -\frac{1}{n}\text{log } P(y^{(1)}, \ldots, y^{(n)} | \tau^{(1)}, \ldots, \tau^{(n)}, \mathbf{w}) =  -\frac{1}{n}\sum_{t=1}^{n} \text{log } P(y^{(t)} | \tau^{(t)}, \mathbf{w})$. Based on Assumption \ref{feedback_assum}, we have $P(y^{(t)} | \tau^{(t)}, \mathbf{w}) = \frac{\text{exp}(\mathbf{w}^T\phi_{y^{(t)}}(\tau^{(t)}))}{\sum^{K-1}_{j=0}\text{exp}(\mathbf{w}^T\phi_j(\tau^{(t)}))}$, which yields

\begin{equation}
    \widehat{\mathbf{w}}_n = \argminB_{\mathbf{w} \in \mathbf{W}_B} -\frac{1}{n}\sum_{t=1}^{n} \text{log } \frac{\text{exp}(\mathbf{w}^T\phi_{y^{(t)}}(\tau^{(t)}))}{\sum^{K-1}_{j=0}\text{exp}(\mathbf{w}^T\phi_j(\tau^{(t)}))}.
    \label{est_west_1}
\end{equation}

Lemma \ref{lemm_concen} bounds the estimation error between the true weight vector and our MLE estimate. This result relies on the fact that the loss function \( \ell^{(n)}(\mathbf{w}) \) is strongly convex in the weight vector \( \mathbf{w} \), and the entire proof is provided in the supplementary material. We use the notation \( \lambda_{\text{min}}(A) \) to denote the smallest eigenvalue of matrix \( A \).

\begin{lemma}
    For any episode \( n \in [N] \), the following holds with probability at least \( 1 - \delta \):

    \begin{equation*}
        \big{|}\big{|}\widehat{\mathbf{w}}_n -\mathbf{w}^\star  \big{|}\big{|}_2  \leq \frac{2}{\eta \lambda_{\min}(\Sigma_{D_n})} \sqrt{\frac{C^2}{2n}\log{\frac{4}{\delta}}}\;, 
    \end{equation*}
 where $ \Sigma_{D_n} = \frac{1}{nK^2} \sum_{i=1}^n \sum_{j=0}^{K-1} \sum_{l=0}^{K-1}(\phi_j(\tau^{(i)})-\phi_l(\tau^{(i)}))$ $(\phi_j(\tau^{(i)})-\phi_l(\tau^{(i)}))^T
$, $ \eta = \frac{\exp(-4B)}{2}$ and $C = \log{\big{(}K\exp(2B)\big{)}}$.
\label{lemm_concen}
\end{lemma}

In this work, we assume that the matrix \( \Sigma_{D_n} \) is well-conditioned and invertible, which implies that \( \lambda_{\text{min}}(\Sigma_{D_n}) \) is bounded away from zero. In practice, one can  use a regularized MLE estimator when the matrix \( \Sigma_{D_n} \) is non-invertible. Lemma \ref{lemm_concen} shows that the \( L_2 \)-norm distance between the MLE estimate \( \widehat{\mathbf{w}}_n \) and the true weight \( \mathbf{w}^{\star} \) remains bounded with high probability. Additionally, the MLE estimate converges to the true weight with high probability for large enough episode numbers \( n \). Since we don't have access to the true weight vector \( \mathbf{w}^{\star} \), we do not have access to the reward function for a trajectory \( R(\tau) \). Instead, using the MLE estimate \( \widehat{\mathbf{w}}_n \), we can construct an estimate of the reward function as follows:

\begin{equation*}
    R(\widehat{\mathbf{w}}_n,\tau) = \sum_{i=0}^{K-1} i\widehat{P}(y_\tau =i) =  \sum_{i=0}^{K-1} i 
 \frac{\text{exp}(\widehat{\mathbf{w}}_n^T\phi_i(\tau))} {\sum^{K-1}_{j=0}\text{exp}(\widehat{\mathbf{w}}_n^{T}\phi_j(\tau))}.
\end{equation*}

Using the concentration result in Lemma \ref{lemm_concen} on the weight parameters, we show the concentration of the estimated reward, that is, \( R(\widehat{\mathbf{w}}_n, \tau) \) converges to the true reward function \( R(\tau) \), as defined in Equation \ref{reward_eq}, with a high probability for sufficiently large \( n \). This result is crucial since the estimated reward function serves as a proxy for the true reward function for the next stage of the algorithm that learns the policy.

\begin{lemma} \label{lem2}
  For any possible trajectory $\tau \in \mathcal{T}$, the following holds with probability at least \( 1 - \delta \):

\begin{equation*}
    \bigg{|}  R(\widehat{\mathbf{w}}_n,\tau) -  R(\tau)\bigg{|} 
    \leq   
\frac{4K\exp(4B)}{\eta \lambda_{\min}(\Sigma_{D_n})} \sqrt{\frac{C^2}{2n}\log{\frac{4}{\delta}}}\;,
\end{equation*}
 where $ \Sigma_{D_n} = \frac{1}{nK^2} \sum_{i=1}^n \sum_{j=0}^{K-1} \sum_{l=0}^{K-1}(\phi_j(\tau^{(i)})-\phi_l(\tau^{(i)}))$ $(\phi_j(\tau^{(i)})-\phi_l(\tau^{(i)}))^T
$, $ \eta = \frac{\exp(-4B)}{2}$ and $C = \log{\big{(}K\exp(2B)\big{)}}$.
\label{reward_concen}
\end{lemma}

\begin{equation}
    \overline{R}(\widehat{\mathbf{w}}_n,\tau) =  \min\left(R(\widehat{\mathbf{w}}_n,\tau) + \frac{4K\exp(4B)}{\eta \lambda_{\min}(\Sigma_{D_n})} \sqrt{\frac{C^2}{2n}\log{\frac{4}{\delta}}},\;K-1\right) \label{Rbar_eq}
\end{equation}

\begin{algorithm}[h]
\hrule
\vspace{0.3em}
\caption{K-UCBVI with Multilevel Episodic Feedback and Known Transition Probabilities.}
\vspace{0.20em}
\hrule
\For{$n=1, \cdots,N $}{

\eIf{$n> 1$}
{
    \begin{equation}
       \pi^{(n)} \in \arg\max_{\pi \in \Pi}  \;\mathbb{E}_{s_0 \sim \rho,\tau \sim \mathbb{P}^{\pi}(.|s_0)} [\overline{R}(\widehat{\mathbf{w}}_n,\tau)]
       \label{equation::pi_t}
    \end{equation}
    
}{
  set $\pi^{(1)}(.| s)$ to be uniform distribution over the action set for all the states.\\
}

 Observe the trajectory $\tau^{(n)} \sim \mathbb{P}^{\pi^{(n)}}$ and the corresponding feedback $y^{(n)}$.\\
 Calculate $\widehat{\mathbf{w}}_n$ by solving Equation~\eqref{est_west_1}.
}

\vspace{0.2em}
\hrule
\vspace{0.3em}
\label{alg_1}
\end{algorithm}

The concentration result in Lemma \ref{reward_concen} shows that the estimate of the reward function using the MLE estimate \( \widehat{\mathbf{w}}_n \) is close to the true reward function. Building upon the principle of optimism in the face of uncertainty---an established framework for online learning algorithms \cite{lattimore2020bandit, chatterji2021theory, faury2020improved}---we utilize this concentration result to define the optimistic reward function, as expressed in Equation \ref{Rbar_eq}, denoted by \( \overline{R}(\widehat{\mathbf{w}}_n, \tau) \). The minimum function in Equation \ref{Rbar_eq} ensures that the optimistic estimate does not go beyond \( K - 1 \), which is the highest level of feedback. By applying Lemma \ref{reward_concen}, we observe that \( R(\tau) \leq \overline{R}(\widehat{\mathbf{w}}_n, \tau) \) with at least \( 1 - \delta \) probability.  It's important to note that the width of the confidence interval in Lemma \ref{reward_concen} scales exponentially with respect to the parameter \( B \), similar to the findings in \cite{chatterji2021theory}. However, in practice, this width can be large due to the constants involved. In our experiments section, to avoid scaling with respect to these constants, we use a confidence interval width of \( \mathcal{O}(\frac{1}{\sqrt{n}}) \).

Our proposed Algorithm \ref{alg_1} involves two key steps: (1) computing the MLE estimate \( \widehat{\mathbf{w}}_n \) using the sampled trajectory and corresponding feedback labels, and (2) computing the policy \( \pi^{(n)} \) by solving the optimization in Equation \ref{equation::pi_t}. Note that the solution to the optimization problem in Equation \ref{equation::pi_t} can generally be a non-Markovian policy, as our reward function depends on the entire trajectory, which may not be Markovian in general. To address this issue, in the next subsection, we propose limiting the search to Markovian policies. This restriction simplifies the problem and allows us to leverage a variety of RL algorithms, including policy gradient methods such as the REINFORCE algorithm—a Monte Carlo variant of policy gradients. By focusing on Markovian policies, we avoid solving the optimization problem in Equation~\ref{equation::pi_t} directly and instead approximate the solution empirically. While one could extend the search to non-Markovian policies by employing memory-based strategies—such as using the entire history, a window of recent history, or a low-dimensional embedding of the history (e.g., through recurrent neural networks)—such approaches often become computationally intractable for Markov Decision Processes (MDPs) with long horizons.

We prove a high-probability regret bound for our proposed Algorithm using the concentration results in Lemma~\ref{reward_concen}, and show that the regret is sublinear in the number of episodes \( N \). The regret bound is stated below, with the detailed regret analysis provided in the supplementary material.

\begin{theorem}
\label{thm1}
   For any \( \delta \in (0,1] \), under Assumptions \ref{feedback_assum} and \ref{assumption::bounded_features}, the cumulative regret of K-UCBVI is upper bounded with probability at least \( 1 - \delta \) as follows:

\begin{align*}
       \mathcal{CR}(N) \leq & \frac{16K\exp(4B)}{\eta \lambda}  \sqrt{\frac{NC^2}{2}\log{\frac{12N}{\delta}}}  +  4K \sqrt{N \log \left(\frac{18N \log N}{\delta}\right)}\;,
\end{align*}
where \( \lambda = \min_{i \in \{1,2,3,\dots,N\}} \lambda_{\min}(\Sigma_{D_n}) \).
\end{theorem}

Theorem~\ref{thm1} shows that the regret of the proposed algorithm scales as \( \mathcal{O}\left(\sqrt{N} \log \frac{N \log N}{\delta}\right) \) with respect to the number of episodes \( N \), with a probability of at least \( 1 - \delta \), which establishes the sublinearity of the regret. The regret scaling of our algorithm with respect to the number of episodes is similar to that in \cite{chatterji2021theory}, which considers binary feedback scores. There is also an exponential dependence on the parameter \( B \), similar to the term \( \kappa \) in the regret bound presented in \cite{chatterji2021theory}. This issue of exponential scaling of regret bounds with respect to the bound parameter has also been discussed in \cite{faury2020improved}.

\subsection{Practical Approximation Using the REINFORCE Algorithm}

In this subsection, we propose an alternative to the optimization in Equation \ref{equation::pi_t}, which is not easy to solve in general. We focus on limiting our search to  Markovian policies where the action depends only on the current state, and we use a REINFORCE-style algorithm \cite{agarwal2019learning} to compute the optimal policy. In general, the deterministic class of policies will not be differentiable, so we can't employ the gradient ascent approach. This motivates us to consider policy classes that are stochastic, which permit differentiability. We restrict ourselves to the class of stochastic Markovian policies parameterized by some $\theta \in \mathbb{R}^d$. Consider a parametric class of policies $\Pi = \{\pi_{\theta}: \theta  \in \mathbb{R}^d\}$. The objective is to maximize the value function over the class of policies, i.e., $\max_{\theta  \in\mathbb{R}^d} V^{\pi_\theta}$. We specifically consider a softmax set of policies (Definition \ref{softmax_pi}), which can model all the stationary (Markovian) policies where the action depends only on the current state. For any realizable trajectory $(\tau \in \mathcal{T})$, we use the notation $Pr^{\pi_{\theta}}_{\rho} (\tau)$ for the probability of realizing trajectory $\tau$ under policy $\pi_{\theta}$ and initial state distribution $\rho$. Also, we have $
{Pr}^{\pi_{\theta}}_{\rho} (\tau) = \rho(s_0) \pi_{\theta}(a_0|s_0) P(s_1|s_0,a_0) \pi_{\theta}(a_1|s_1) \ldots P(s_H|s_{H-1},a_{H-1}).$

\begin{definition} [Softmax Policies] The softmax policy is an explicit tabular representation of a policy defined as:

\begin{equation*}
    \pi_{\theta}(a|s) = \frac{\exp{\theta_{a,s}}}{\sum_{a' \in \mathcal{A}}\exp{\theta_{a',s}}}. 
\end{equation*}

\label{softmax_pi}
\end{definition}

\vspace{0.2em}

The parameter space is $\Theta \in \mathbb{R}^{|\mathcal{S}||\mathcal{A}|}$. The closure of the set of softmax policies contains all stationary and stochastic policies. Instead of the optimization problem in Equation \ref{equation::pi_t}, we solve
$$
\arg \max_{\theta \in \mathbb{R}^{|\mathcal{S}||\mathcal{A}|}} \overline{V}_n^{\pi_{\theta}}: = \mathbb{E}_{\tau \sim  Pr^{\pi_{\theta}}_{\rho}} \left[ \overline{R}(\widehat{\mathbf{w}}_n, \tau) \right]
$$

$\overline{V}_n^{\pi_{\theta}}$ is the optimistic value function at episode $n$. The gradient of the optimistic value function of policy $\pi_{\theta}$ with respect to the parameter vector $\theta$ is given by:

\begin{equation*}
     \nabla \overline{V}^{\pi_{\theta}}_{n} = \mathbb{E}_{\tau \sim  Pr^{\pi_{\theta}}_{\rho}} \left[  \overline{R}(\widehat{\mathbf{w}}_n, \tau)  \sum_{t=0}^{H} \nabla \log(\pi_{\theta}(a_t|s_t)) \right].
\end{equation*}

This follows from the fact that
$
\nabla  \overline{V}^{\pi_{\theta}}_{n} = \sum_{\tau \in \mathcal{T}} \overline{R}(\widehat{\mathbf{w}}_n, \tau) Pr^{\pi_{\theta}}_{\rho} (\tau) \nabla \log Pr^{\pi_{\theta}}_{\rho} (\tau).
$
Using the definition of $Pr^{\pi_{\theta}}_{\rho} (\tau)$, we obtain the final result
$
\nabla  \overline{V}^{\pi_{\theta}}_{n} = \mathbb{E}_{\tau \sim  Pr^{\pi_{\theta}}_{\rho}} \left[\overline{R}(\widehat{\mathbf{w}}_n, \tau) \sum_{t=0}^{H} \nabla \log(\pi_{\theta}(a_t|s_t)) \right].
$

\begin{remark} For the softmax class of policies where $\pi_{\theta}(a_t|s_t) = \frac{\exp{\theta_{a_t,s_t}}}{\sum_{a' \in \mathcal{A}}\exp{\theta_{a_t',s_t}}}$, the partial derivatives  $\frac{\partial \log(\pi_{\theta}(a_t|s_t))}{\partial \theta_{a,s}} = \mathbbm{1}_{s_t=s} [\mathbbm{1}_{a_t=a} - \pi_{\theta}(a_t|s_t)]$. This property allows the gradient $\nabla \log(\pi_{\theta_t}(a_t|s_t)$ to be computed efficiently.
\end{remark}

In this practical implementation, the objective is to maximize the optimistic value function with respect to the parameter vector $\theta$, which parameterizes the policy. We use the stochastic gradient ascent algorithm for this purpose. At each step $n$ of gradient ascent, we sample a fixed number of trajectories using the transition probabilities and the policy $\pi_{\theta_i}$. Then, we compute the average policy gradient over these samples and update the parameter vector as $\theta_{i+1} = \theta_{i} + \eta \nabla \overline{V}^{\pi_{\theta_i}}_{n}$ until some termination criteria are met, such as when the change in the parameter vector becomes small.

\section{EXPERIMENTS}
This section presents the experiments conducted to evaluate the optimistic learning algorithm from multi-level feedback. We test the proposed algorithm on various grid-world environments and demonstrate empirically that the agent is able to learn an optimal policy from multi-level feedback. Specifically, we showcase the results from an $8\times8$ grid-world environment with the feedback level $K=4$ and $6$. The practical implementation of the algorithm, as used in the experiments, is detailed in Algorithm 2 of the supplementary material.

\subsection{Simulation Setting}\label{subsec:5.1}
\begin{figure*}[h!]
\centering

\begin{minipage}[t]{0.3\textwidth} 
    \centering
    \includegraphics[height=2.8cm, width=4.1cm]{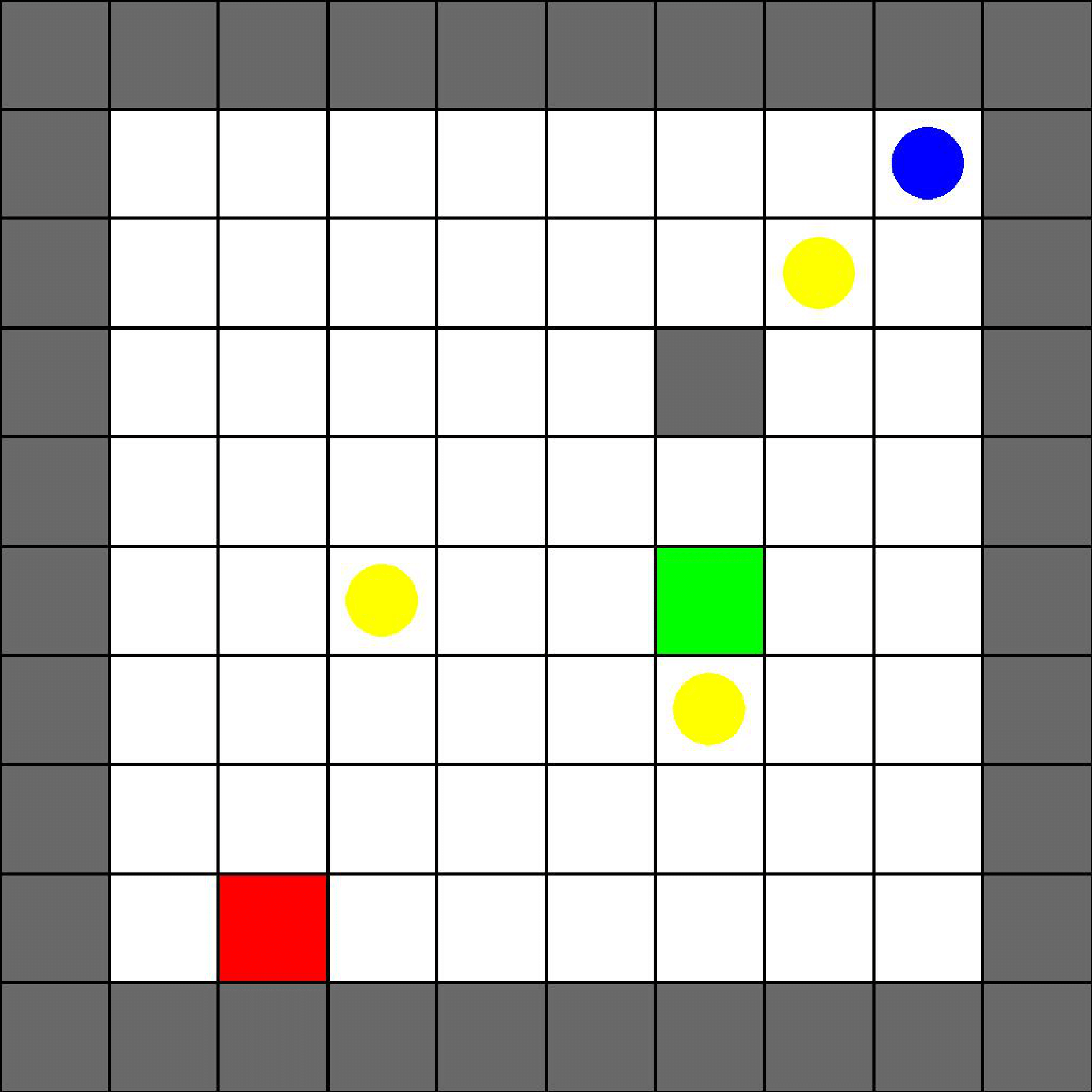}  
    \textbf{(a)}
\end{minipage}
\hspace{1em} 
\begin{minipage}[t]{0.3\textwidth}
    \centering
    \includegraphics[height=3.1cm, width=5cm]{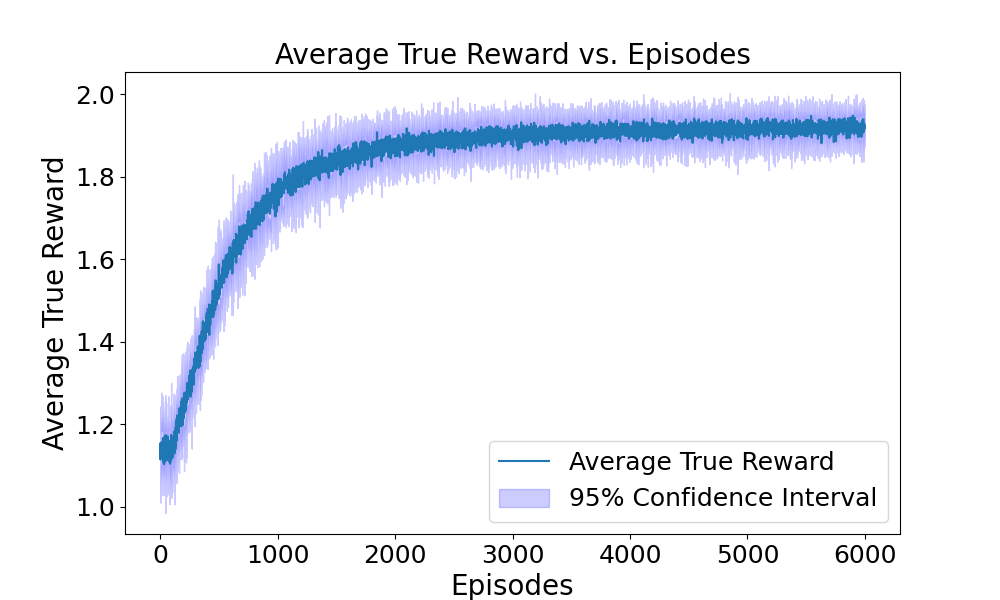}
    \textbf{(b)}
\end{minipage}
\hspace{1em} 
\begin{minipage}[t]{0.3\textwidth}
    \centering
    \includegraphics[height=3.1cm, width=5cm]{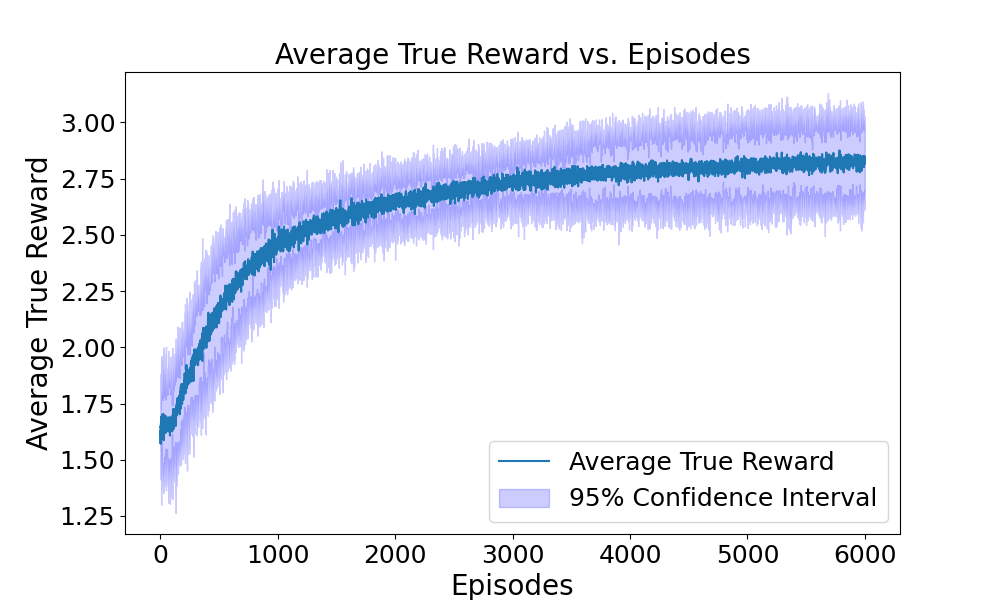}
    \textbf{(c)}
\end{minipage}
\caption{(a) $8\times8$ grid-world environment with the danger state (red cell), wall state (gray cells) and goal state (green cell) depicted. (b) Plot of the average true reward against the number of episodes for $K=4$. (c) Plot of the average true reward against the number of episodes for $K=6$.}
 \label{fig:env}
\end{figure*}

We describe the simulation setup for an $8\times8$ grid-world depicted in Figure \ref{fig:env}(a) with the given state configurations. The goal of the agent, represented with the blue circle, is to collect the coins indicated by the three yellow circles and then reach the goal state depicted by the green square. In addition, the agent should avoid the absorbing danger state represented by the red square where feedback is the lowest. Walls are represented as gray cells. We set the horizon $H = 50$. The agent can move to any cell on the grid-world with actions—up, down, left, or right. The agent can move to the intended cell with a probability $0.91$ and to the remaining three cells with a probability of 0.03 each.

\vspace{-0.5em}

The true reward model, parameterized by $\mathbf{w}^\star$ is unknown to the agent. To generate a realistic $\mathbf{w}^\star$, we design a rule-based reward mechanism and learn a corresponding $\mathbf{w}^\star$ that defines how well the agent achieves the goal. Particularly, for any feedback level, we assign a reward of $0$ if the agent reaches the danger state at the end of the episode. Otherwise, the agent gets an additional level of feedback for each collected coin with the highest feedback assigned if all coins have been collected and the agent reached the goal. We define the features $\phi(\tau)$ as a vector with dimension $(4+c)$ where $c$ is the number of coins. The elements of the vector are the Manhattan distance between the final state at the end of the horizon $s^\tau_H$ and the goal state G, the Manhattan distance between $s^\tau_H$ and the danger state, an indicator function that represents whether or not the agent is at the goal state, an indicator function that represents whether or not the agent is at the danger state, and an indicator function for each coin determining whether or not it was picked. Given a softmax policy $\pi_\theta$ parameterized by $\theta$, we train and update $\theta$ using the REINFORCE algorithm. Empirically, we approximate the expectation in the gradient of the value function by sampling 50 trajectories. At convergence, we sample a trajectory and the corresponding feedback from $\mathbf{w}^\star$ and update $\widehat{\mathbf{w}}$ using the projected gradient descent step.  We repeat the experiment over $20$ independent runs and $6000$ episodes. The plots of the average true reward against the number of episodes with a confidence interval of $\pm2$ standard deviations are presented in Figures \ref{fig:env}(a) and (b).


\subsection{Simulation Results}
\begin{figure*}[h!]
\centering

\begin{minipage}[t]{0.3\textwidth} 
    \centering
    \includegraphics[height=3.1cm, width=5.2cm]{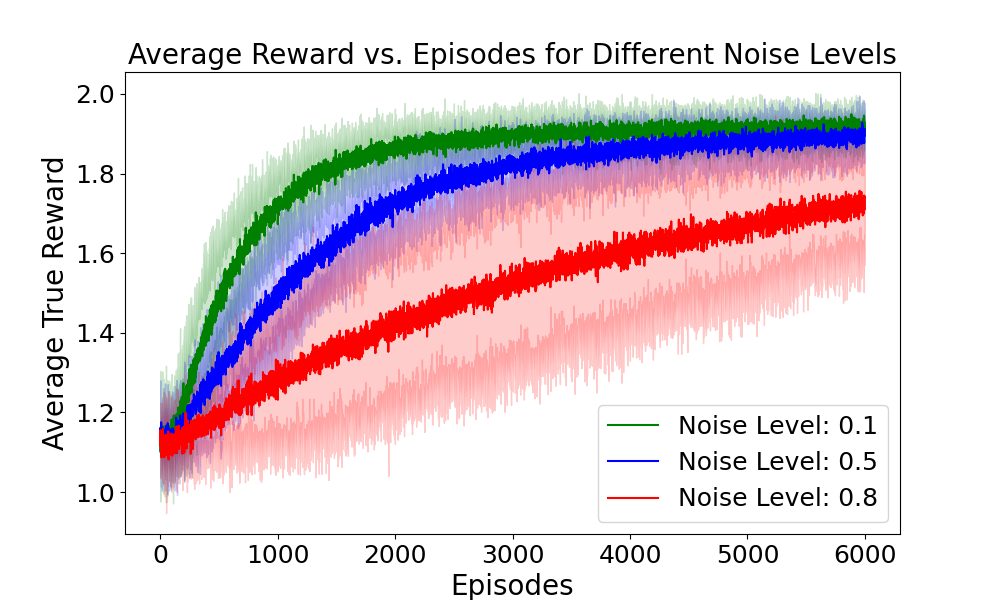}  
    \textbf{(a)}
\end{minipage}
\hspace{1em} 
\begin{minipage}[t]{0.3\textwidth}
    \centering
    \includegraphics[height=3.1cm, width=5.2cm]{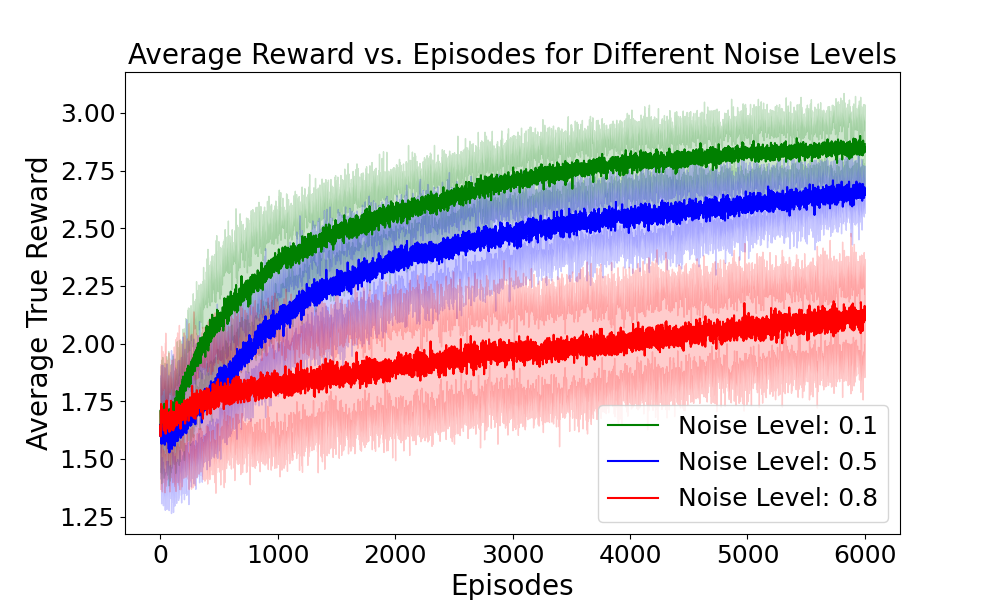}
    \textbf{(b)}
\end{minipage}
\hspace{1em}
\begin{minipage}[t]{0.3\textwidth}
    \centering
    \includegraphics[height=3.1cm, width=5.2cm]{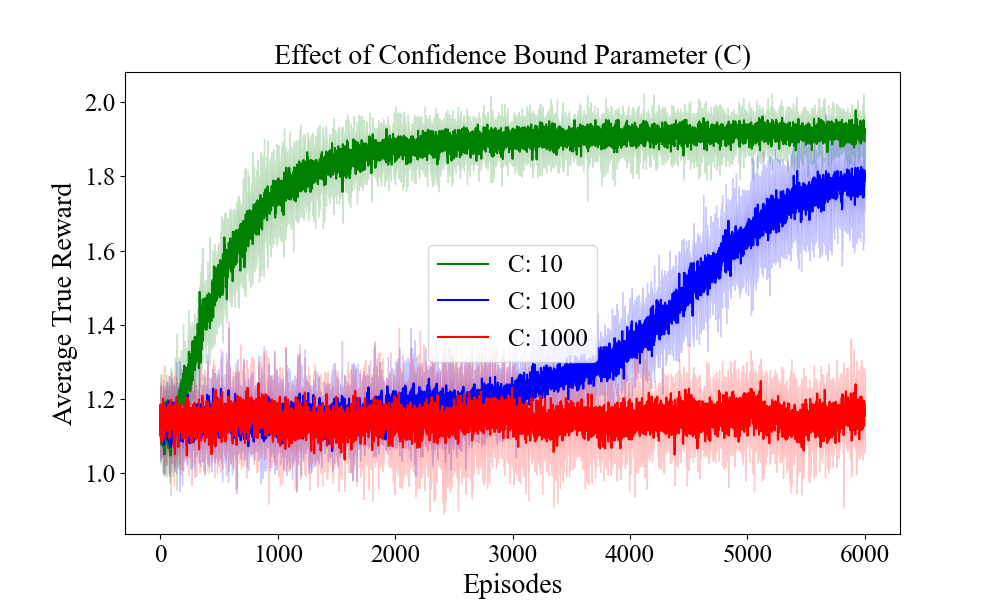}
    \textbf{(c)}
\end{minipage}
\hspace{1em}
\caption{(a) Impact of varying noisy feedback on the learned policy $(K=4)$. (b) Impact of varying noisy feedback on the learned policy $(K=6)$. (c) Effect of the confidence bound parameter on the learned policy $(K=4)$.}
 \label{fig:comparison_plots}
\end{figure*}

Human feedback is inherently prone to bias, and preference models like the Bradley-Terry model \cite{bradley1952rank} and the categorical model proposed in this work, may not accurately represent the feedback. Hence, we empirically analyze the robustness of the proposed algorithm to model misspecification. To do this, we introduce perturbations in the form of noise modeled as a uniform distribution over the feedback levels. We define the noisy feedback as a convex combination of the true feedback and varying levels of noise. Figure \ref{fig:comparison_plots}(a) and (b) illustrate the impact of different noise levels on the performance of the learned policy. For instance, at a noise level of $0.1$, $10\%$ of the feedback is sampled from the noise distribution, while $90\%$ is sampled from the true feedback.
It can be observed that incorporating noise affects the rate of convergence of the policy with slower convergence rates as the noise level increases.

As stated in section \ref{sec:4}, it should be noted that in the experiments, we set the confidence term in the optimistic reward function to $(C\times \frac{1}{\sqrt{n}})$ where $C$ is confidence bound parameter and $n$ is the number of samples or episodes. For our experiments, we set $C = 10$. Figure \ref{fig:comparison_plots}(c) compares the effects of different values of $C$ on the learned policy.

\vspace{-1em}

\section{CONCLUSION AND FUTURE WORK}
In this paper, we consider a multi-level feedback mechanism where the agent receives a score at the end of each episode. We design an optimistic algorithm for learning an optimal policy from these feedback scores and prove that, under mild assumptions, the algorithm achieves cumulative regret that scales sublinearly with respect to the number of episodes. We perform experiments on various grid-world environments to demonstrate that the algorithm learns an optimal policy. Furthermore, we conduct perturbation analysis on the feedback model to assess the impact of noise on the performance of the policy. As future work, we propose to build upon our quantized feedback model—which discretizes continuous reward signals into finite set of levels—by exploring information-theoretic approaches to understand how such quantization can affect the optimality of the learned policy. Additionally, we aim to investigate alternative preference models and incorporate more robust analysis techniques to address challenges such as model misspecification.

\section{ACKNOWLEDGMENTS}
This work was supported in part by the National Aeronautics and Space Administration under Grant 80NSSC24M0070.


\bibliography{sample_paper.bib}

\clearpage
\appendix
\section{Supplementary Material} 

In the following subsections, we provide comprehensive and formal mathematical proofs for the theoretical results in the main paper.

\subsection{Proof of Lemma \ref{lemm_concen}}
We recall that the maximum likelihood estimate (MLE) \( \widehat{\mathbf{w}}_n \) based on the sampled trajectories \( \{\tau^{(i)}\}_{i=1}^{n} \) and the corresponding categorical feedback received by \( \{y^{(i)}\}_{i=1}^{n} \) is defined as follows:

\begin{equation*}
    \widehat{\mathbf{w}}_n = \argminB_{\mathbf{w} \in \mathbf{W}_B}  \ell^{(n)}(\mathbf{w}) = \argminB_{\mathbf{w} \in \mathbf{W}_B} -\frac{1}{n}\sum_{i=1}^{n} \text{log } \frac{\text{exp}(\mathbf{w}^T\phi_{y^{(i)}}(\tau^{(i)}))}{\sum^{K-1}_{j=0}\text{exp}(\mathbf{w}^T\phi_j(\tau^{(i)}))}
    \label{est_west}
\end{equation*}
We compute the gradient of loss function $\ell^{(n)}(\mathbf{w})$ as follows:

$$\nabla\ell^{(n)}(\mathbf{w}) = - \frac{1}{n}\sum_{i=1}^n\sum_{j=0}^{K-1} \frac{\text{exp}(\mathbf{w}^T\phi_j(\tau^{(i)}))}{\sum_{l=0}^{K-1}\text{exp}(\mathbf{w}^T\phi_l(\tau^{(i)}))}\bigg(\phi_{y^{(i)}}(\tau^{(i)})-\phi_j(\tau^{(i)})\bigg)$$
In order to compute the Hessian matrix, we start by calculating $ \frac{\delta}{\delta w[\alpha]} \nabla \ell^{(n)}(\mathbf{w}) $, where $ w[\alpha] $ is the $\alpha$-th component of the vector $ \mathbf{w} $. Also, note that the $\alpha$-th row of $ \nabla^2 \ell^{(n)}(\mathbf{w}) $ is given by $ \frac{\delta}{\delta w[\alpha]} \nabla \ell^{(n)}(\mathbf{w}) $.

\begin{align}
\alpha^{\text{th}} & \text{ row of Hessian Matrix } (\nabla^2 \ell^{(n)}(\mathbf{w})) = \nonumber \\
&\frac{-1}{n} \sum_{i=1}^{n} \sum_{j=0}^{K-1} \frac{\sum_{l=0}^{K-1} \text{exp}(\mathbf{w}^T (\phi_j(\tau^{(i)})+\phi_l(\tau^{(i)}))) \; (\phi_{j}(\tau^{(i)})[\alpha]- \phi_{l}(\tau^{(i)})[\alpha]) }{\left(\sum_{l=0}^{K-1} \text{exp}(\mathbf{w}^T \phi_l(\tau^{(i)}))\right)^2} \bigg( M_{ij}\bigg)
\label{eq_pff}
\end{align}
Where $M_{ij} = \phi_{y^{(i)}}(\tau^{(i)}) - \phi_j(\tau^{(i)})$. The Hessian matrix can be computed by combining the rows obtained from Equation \ref{eq_pff} as follows:

\begin{equation*}
\nabla^2 \ell^{(n)}(\mathbf{w}) =\frac{1}{n} \sum_{i=1}^n\sum_{j=0}^{K-1} \sum_{l=0}^{K-1} \frac{\text{exp}(\mathbf{w}^T(\phi_j(\tau^{(i)})+\phi_l(\tau^{(i)})))}{(\sum_{l=0}^{K-1}\text{exp}(\mathbf{w}^T\phi_l(\tau^{(i)})))^2}(\phi_j(\tau^{(i)})-\phi_l(\tau^{(i)}))(\phi_{j}(\tau^{(i)})-\phi_{y^{(i)}}(\tau^{(i)}))^T    
\end{equation*}

\begin{multline*}
 \nabla^2 \ell^{(n)}(\mathbf{w}) =\frac{1}{n} \sum_{i=1}^n\sum_{j=0}^{K-1} \sum_{l=0}^{K-1} \frac{\text{exp}(\mathbf{w}^T(\phi_j(\tau^{(i)})+\phi_l(\tau^{(i)})))}{(\sum_{l=0}^{K-1}\text{exp}(\mathbf{w}^T\phi_l(\tau^{(i)})))^2}(\phi_j(\tau^{(i)})-\phi_l(\tau^{(i)}))(\phi_{j}(\tau^{(i)})-\phi_{l}(\tau^{(i)}))^T   \\ - \frac{1}{n} \sum_{i=1}^n\sum_{j=0}^{K-1} \sum_{l=0}^{K-1} \frac{\text{exp}(\mathbf{w}^T(\phi_j(\tau^{(i)})+\phi_l(\tau^{(i)})))}{(\sum_{l=0}^{K-1}\text{exp}(\mathbf{w}^T\phi_l(\tau^{(i)})))^2}(\phi_j(\tau^{(i)})-\phi_l(\tau^{(i)}))(\phi_{y^{(i)}}(\tau^{(i)})-\phi_{l}(\tau^{(i)}))^T   
\end{multline*}

\begin{multline*}
 \nabla^2 \ell^{(n)}(\mathbf{w}) =\frac{1}{n} \sum_{i=1}^n\sum_{j=0}^{K-1} \sum_{l=0}^{K-1} \frac{\text{exp}(\mathbf{w}^T(\phi_j(\tau^{(i)})+\phi_l(\tau^{(i)})))}{(\sum_{l=0}^{K-1}\text{exp}(\mathbf{w}^T\phi_l(\tau^{(i)})))^2}(\phi_j(\tau^{(i)})-\phi_l(\tau^{(i)}))(\phi_{j}(\tau^{(i)})-\phi_{l}(\tau^{(i)}))^T   \\ - \frac{1}{n} \sum_{i=1}^n \sum_{j=0}^{K-1} \sum_{l=0}^{K-1} \frac{\text{exp}(\mathbf{w}^T(\phi_j(\tau^{(i)})+\phi_l(\tau^{(i)})))}{(\sum_{l=0}^{K-1}\text{exp}(\mathbf{w}^T\phi_l(\tau^{(i)})))^2}  
 (\phi_j(\tau^{(i)})-\phi_l(\tau^{(i)}))(\phi_{y^{(i)}}(\tau^{(i)})-\phi_{l}(\tau^{(i)}))^T   
\end{multline*}

\begin{multline*}
 \nabla^2 \ell^{(n)}(\mathbf{w}) =\frac{1}{n} \sum_{i=1}^n\sum_{j=0}^{K-1} \sum_{l=0}^{K-1} \frac{\text{exp}(\mathbf{w}^T(\phi_j(\tau^{(i)})+\phi_l(\tau^{(i)})))}{(\sum_{l=0}^{K-1}\text{exp}(\mathbf{w}^T\phi_l(\tau^{(i)})))^2}(\phi_j(\tau^{(i)})-\phi_l(\tau^{(i)}))(\phi_{j}(\tau^{(i)})-\phi_{l}(\tau^{(i)}))^T   \\ - \frac{1}{n} \sum_{i=1}^n 
 \sum_{l=0}^{K-1} \sum_{j=0}^{K-1}  
\frac{\text{exp}(\mathbf{w}^T(\phi_l(\tau^{(i)})+\phi_j(\tau^{(i)})))}{(\sum_{l=0}^{K-1}\text{exp}(\mathbf{w}^T\phi_l(\tau^{(i)})))^2} 
  (\phi_l(\tau^{(i)})-\phi_j(\tau^{(i)}))(\phi_{l}(\tau^{(i)}) - \phi_{y^{(i)}}(\tau^{(i)}))^T   
\end{multline*}

\begin{multline*}
 \nabla^2 \ell^{(n)}(\mathbf{w}) =\frac{1}{n} \sum_{i=1}^n\sum_{j=0}^{K-1} \sum_{l=0}^{K-1} \frac{\text{exp}(\mathbf{w}^T(\phi_j(\tau^{(i)})+\phi_l(\tau^{(i)})))}{(\sum_{l=0}^{K-1}\text{exp}(\mathbf{w}^T\phi_l(\tau^{(i)})))^2}(\phi_j(\tau^{(i)})-\phi_l(\tau^{(i)}))(\phi_{j}(\tau^{(i)})-\phi_{l}(\tau^{(i)}))^T  \\ -    \nabla^2 \ell^{(n)}(\mathbf{w}) 
\end{multline*}

\begin{multline*}
 \nabla^2 \ell^{(n)}(\mathbf{w}) =\frac{1}{n} \sum_{i=1}^n\sum_{j=0}^{K-1} \sum_{l=0}^{K-1} \frac{\text{exp}(\mathbf{w}^T(\phi_j(\tau^{(i)})+\phi_l(\tau^{(i)})))}{2(\sum_{l=0}^{K-1}\text{exp}(\mathbf{w}^T\phi_l(\tau^{(i)})))^2}(\phi_j(\tau^{(i)})-\phi_l(\tau^{(i)}))(\phi_{j}(\tau^{(i)})-\phi_{l}(\tau^{(i)}))^T  
\end{multline*}

\textbf{Strong convexity of $\ell^{(n)}(\mathbf{w})$}. We want to show that $\ell^{(n)}(\mathbf{w})$ is strongly convex at $\mathbf{w}^{\star}$ with respect to the semi-norm $\|\cdot\|_{\Sigma_{\mathcal{D}}}$ which implies there is some constant $\eta>0$ such that:
$$
\ell^{(n)}\left(\mathbf{w}^{*}+\Delta\right)-\ell^{(n)}\left(\mathbf{w}^{*}\right)-\left\langle\nabla \ell^{(n)}\left(\mathbf{w}^{*}\right), \Delta\right\rangle \geq \eta\|\Delta\|_{\Sigma_{\mathcal{D}}}^2
$$
for all perturbations $\Delta \in \mathbb{R}^d$ such that $\mathbf{w}^{*} + \Delta \in \mathbf{W}_B$, we can establish the strong convexity by showing that 
$v^T \nabla^2 \ell^{(n)}(\mathbf{w}) v \geq \eta \|v\|_A^2$
where $A$ is some PD matrix and $\eta \geq 0$. Since we assume bounded feature vectors and the true weight vector in Assumption \ref{assumption::bounded_features}, we have 
$\frac{w^T (\phi_j(\tau^{(i)}) + \phi_l(\tau^{(i)}))}{2 \left( \sum_{l=0}^{K-1} \exp(\mathbf{w}^T \phi_l(\tau^{(i)})) \right)^2} \geq \frac{\exp(-4B)}{2K^2}$.
We set
$\eta = \frac{\exp(-4B)}{2}$.

\begin{equation*}
    v^T\nabla^2\ell^{(n)}(\mathbf{w})v \geq \frac{\eta } {nK^2} v^T\bigg( \sum_{i=1}^n\sum_{j=0}^{K-1} \sum_{l=0}^{K-1} (\phi_j(\tau^{(i)})-\phi_l(\tau^{(i)}))(\phi_{j}(\tau^{(i)})-\phi_{l}(\tau^{(i)}))^T \bigg{)} v
\end{equation*}

We define: $\Sigma_{D_n} = \frac{1}{nK^2} \sum_{i=1}^n \sum_{j=0}^{K-1} \sum_{l=0}^{K-1}(\phi_j(\tau^{(i)})-\phi_l(\tau^{(i)}))(\phi_j(\tau^{(i)})-\phi_l(\tau^{(i)}))^T$.

$$v^T\nabla^2\ell^{(n)}(\mathbf{w})v \geq \eta v^T \Sigma_{D_n} v
=\eta||v||_{\Sigma_{D_n}}^2$$
Hence, we have shown that $\ell^{(n)}(\mathbf{w})$ is strongly convex with respect to the semi-norm $\|\cdot\|_{\Sigma_{D_n}}$. As mentioned in the main paper, we assume that the matrix $\Sigma_{D_n}$ is invertible. \\

$$\ell^{(n)}(\mathbf{w}) = \frac{1}{n}\sum_{i=1}^n -\text{log}P(y^{(i)}|\tau^{(i)},w) = - \frac{1}{n}\sum_{i=1}^n\text{log}\frac{\text{exp}(\mathbf{w}^T\phi_{y^{(i)}}(\tau^{(i)}))}{\sum^{K-1}_{j=0}\text{exp}(\mathbf{w}^T\phi_j(\tau^{(i)}))} $$
Let us define:
\begin{equation*}
    S_i = -\text{log}\frac{\text{exp}(\mathbf{w}^T\phi_{y^{(i)}}(\tau^{(i)}))}{\sum^{K-1}_{j=0}\text{exp}(\mathbf{w}^T\phi_j(\tau^{(i)}))}
\end{equation*}

$$\ell^{(n)}(\mathbf{w}) = \frac{1}{n}\sum_{i=1}^n -\text{log}P(y^{(i)}|\tau^{(i)},w) = - \frac{1}{n}\sum_{i=1}^n S_i $$
Note that $0 \leq S_i \leq -\log{\exp{\left(\frac{-2B}{K}\right)}} = \log{\left(K \exp{(2B)}\right)}$. We define  $C := \log{\left(K \exp{(2B)}\right)}$. We denote the true loss function by $\ell^*(\mathbf{w}) = \mathbb{E}[S_i]$, where the expectation is with respect to the randomly sampled trajectories and corresponding feedback labels. Using Hoeffding's inequality, we have:

\begin{equation}
    |\ell^{(n)}(\mathbf{w}) - \ell^*(\mathbf{w})| \leq \sqrt{\frac{C^2}{2n}\log{\frac{2}{\delta}}} \quad \text{w.p. at least} \;\; 1-\delta \label{hoefeq}
\end{equation}
The true weight vector $\mathbf{w}^{*}$ is the minimizer of the true loss function, and $\widehat{\mathbf{w}}_n$ is the minimizer of $\ell^{(n)}(\mathbf{w})$. We have the following:

\begin{equation*}
    \ell^{*}(\widehat{\mathbf{w}}_n) - \sqrt{\frac{C^2}{2n}\log{\frac{2}{\delta}}} \leq  \ell^{(n)}(\widehat{\mathbf{w}}_n) \leq \ell^{*}(\widehat{\mathbf{w}}_n) + \sqrt{\frac{C^2}{2n}\log{\frac{2}{\delta}}}  \quad \text{w.p. at least} \;\; 1-\delta
\end{equation*}

\begin{equation*}
    \ell^{*}(\mathbf{w}^{*}) - \sqrt{\frac{C^2}{2n}\log{\frac{2}{\delta}}} \leq  \ell^{(n)}(\mathbf{w}^{*}) \leq \ell^{*}(\mathbf{w}^{*}) + \sqrt{\frac{C^2}{2n}\log{\frac{2}{\delta}}}  \quad \text{w.p. at least} \;\; 1-\delta
\end{equation*}
Combining the above equations and using union bound we have the following:

\begin{equation*}
   |\ell^{(n)}(\mathbf{w}^{*})-\ell^{(n)}(\widehat{\mathbf{w}}_n)| \leq    |\ell^{*}(\widehat{\mathbf{w}}_n)-\ell^{*}(\mathbf{w}^{*})| + 2 \sqrt{\frac{C^2}{2n}\log{\frac{2}{\delta}}}  \quad \text{w.p. at least} \;\; 1-2\delta
\end{equation*}
Similarly we can show that:

\begin{equation*}
|\ell^{*}(\widehat{\mathbf{w}}_n)-\ell^{*}(\mathbf{w}^{*})|\leq    |\ell^{(n)}(\mathbf{w}^{*})-\ell^{(n)}(\widehat{\mathbf{w}}_n)| + 2 \sqrt{\frac{C^2}{2n}\log{\frac{2}{\delta}}}  \quad \text{w.p. at least} \;\; 1-2\delta
\end{equation*}
By the strong convexity of the loss function, we have that:
\begin{equation*}
    \ell^{(n)}(\mathbf{y}) - \ell^{(n)}(\mathbf{x}) + \nabla \ell^{(n)}(x)^T(\mathbf{y}-\mathbf{x}) \geq \eta ||\mathbf{y}-\mathbf{x}||_{\Sigma_{D_n}}
\end{equation*}
We define $\Delta = \widehat{\mathbf{w}}_n - \mathbf{w}^{*}$. Let $\mathbf{x} = \widehat{\mathbf{w}}_n$ and $\mathbf{y} = \mathbf{w}^{*} = \widehat{\mathbf{w}}_n - \Delta$. 

\begin{equation*}
    \ell^{(n)}( \widehat{\mathbf{w}}_n - \Delta) - \ell^{(n)}(\widehat{\mathbf{w}}_n) + \nabla \ell^{(n)}(\widehat{\mathbf{w}}_n)^T(-\Delta) \geq \eta ||\Delta||_{\Sigma_{D_n}}
\end{equation*}
Since $\widehat{\mathbf{w}}_n$ is the minimizer of the loss function which is strongly convex function the gradient vector $\nabla\ell^{(n)}(\widehat{\mathbf{w}}_n)$ is in fact a zero vector.

\begin{equation*}
  \eta ||\Delta||_{\Sigma_{D_n}}  \leq  \ell^{(n)}( \widehat{\mathbf{w}}_n - \Delta) - \ell^{(n)}(\widehat{\mathbf{w}}_n) 
\end{equation*}

\begin{equation*}
  \eta ||\Delta||_{\Sigma_{D_n}}  \leq  \ell^{(n)}( \mathbf{w}^{*}) - \ell^{(n)}(\widehat{\mathbf{w}}_n) 
\end{equation*}
Since $\mathbf{w}^{*}$ is the minimizer of the true loss function we have the following:

\begin{align*}
      \eta ||\Delta||_{\Sigma_{D_n}}  &\leq  \ell^{(n)}( \mathbf{w}^{*}) - \ell^{(n)}(\widehat{\mathbf{w}}_n)  + \ell^{*}(\widehat{\mathbf{w}}_n) - \ell^{*}(\mathbf{w}^*)\\
       &  \leq  \bigg{(}\ell^{(n)}( \mathbf{w}^{*}) - \ell^{*}(\mathbf{w}^*)\bigg{)} + \bigg{(}\ell^{*}(\widehat{\mathbf{w}}_n)-\ell^{(n)}(\widehat{\mathbf{w}}_n)  \bigg{)}   
\end{align*}

From the result in equation \ref{hoefeq} we have the following:

\begin{align*}
      \eta ||\Delta||_{\Sigma_{D_n}}  &\leq  \sqrt{\frac{C^2}{2n}\log{\frac{2}{\delta}}} + \sqrt{\frac{C^2}{2n}\log{\frac{2}{\delta}}} \quad \text{w.p. at least} \;\; 1-2\delta\\
      &  \leq 2 \sqrt{\frac{C^2}{2n}\log{\frac{2}{\delta}}} \quad \text{w.p. at least} \;\; 1-2\delta 
\end{align*}

\begin{equation*}
 ||\widehat{\mathbf{w}}_n-\mathbf{w}^{*}||_{\Sigma_{D_n}}  \leq \frac{2}{\eta} \sqrt{\frac{C^2}{2n}\log{\frac{2}{\delta}}} \quad \text{w.p. at least} \;\; 1-2\delta 
\end{equation*}
Equivalently, we have the following high-probability concentration bound:

\begin{equation*}
||\widehat{\mathbf{w}}_n-\mathbf{w}^{*}||_2  \leq \frac{2}{\eta \lambda_{\min}(\Sigma_{D_n})} \sqrt{\frac{C^2}{2n}\log{\frac{2}{\delta}}} \quad \text{w.p. at least} \;\; 1-\delta
\end{equation*}

where 
\[
\Sigma_{D_n} = \frac{1}{nK^2} \sum_{i=1}^n \sum_{j=0}^{K-1} \sum_{l=0}^{K-1}(\phi_j(\tau^{(i)})-\phi_l(\tau^{(i)}))(\phi_j(\tau^{(i)})-\phi_l(\tau^{(i)}))^T,
\]
\[
\eta = \frac{\exp(-4B)}{2}, \quad \text{and} \quad C = \log{(K \exp(2B))}.
\]
This completes the Proof of the Lemma \ref{lemm_concen}.

\subsection{Proof of Lemma \ref{reward_concen}}

The true reward and the estimated reward for any trajectory $\tau \in \mathcal{T}$ are defined respectively as

\begin{equation*}
    R(\tau) = \sum_{i=0}^{K-1} iP(y_\tau =i) =  \sum_{i=0}^{K-1} i 
 \frac{\text{exp}(\mathbf{w}^{*T}\phi_i(\tau))} {\sum^{K-1}_{j=0}\text{exp}(\mathbf{w}^{*T}\phi_j(\tau))},
\end{equation*}

\begin{equation*}
    R(\widehat{\mathbf{w}}_n,\tau) = \sum_{i=0}^{K-1} i\widehat{P}(y_\tau =i) =  \sum_{i=0}^{K-1} i 
 \frac{\text{exp}(\widehat{\mathbf{w}}_n^{T}\phi_i(\tau))} {\sum^{K-1}_{j=0}\text{exp}(\widehat{\mathbf{w}}_n^{T}\phi_j(\tau))}.
\end{equation*}

\begin{align*}
     \big{|}  R(\widehat{\mathbf{w}}_n,\tau) -  R(\tau)\big{|} &= \bigg{|}\sum_{i=0}^{K-1} i 
 \frac{\text{exp}(\mathbf{w}^{*T}\phi_i(\tau))} {\sum^{K-1}_{j=0}\text{exp}(\mathbf{w}^{*T}\phi_j(\tau))} -  \sum_{i=0}^{K-1} i 
 \frac{\text{exp}(\widehat{\mathbf{w}}_n^{T}\phi_i(\tau))} {\sum_{i=0}^{K-1}\text{exp}(\widehat{\mathbf{w}}_n^{T}\phi_j(\tau))}\bigg{|}\\
& = \bigg{|}\sum_{i=0}^{K-1} i \bigg{(}  \frac{\text{exp}(\mathbf{w}^{*T}\phi_i(\tau))} {\sum^{K-1}_{j=0}\text{exp}(\mathbf{w}^{*T}\phi_j(\tau))} -   \frac{\text{exp}(\widehat{\mathbf{w}}_n^{T}\phi_i(\tau))} {\sum^{K-1}_{j=0}\text{exp}(\widehat{\mathbf{w}}_n^{T}\phi_j(\tau))}\bigg{)}
  \bigg{|}\\
  &  \leq  \sum_{i=0}^{K-1} i \bigg{|} \frac{\text{exp}(\mathbf{w}^{*T}\phi_i(\tau))} {\sum^{K-1}_{j=0}\text{exp}(\mathbf{w}^{*T}\phi_j(\tau))} -   \frac{\text{exp}(\widehat{\mathbf{w}}_n^{T}\phi_i(\tau))} {\sum^{K-1}_{j=0}\text{exp}(\widehat{\mathbf{w}}_n^{T}\phi_j(\tau))} \bigg{|}\\
  &
    \leq  K \sum_{i=0}^{K-1}\bigg{|} \frac{\text{exp}(\mathbf{w}^{*T}\phi_i(\tau))} {\sum^{K-1}_{j=0}\text{exp}(\mathbf{w}^{*T}\phi_j(\tau))} -   \frac{\text{exp}(\widehat{\mathbf{w}}_n^{T}\phi_i(\tau))} {\sum^{K-1}_{j=0}\text{exp}(\widehat{\mathbf{w}}_n^{T}\phi_j(\tau))} \bigg{|}
\end{align*}
The first inequality holds due to triangular inequality and last inequality holds due to the fact $i < K$. We define $f_{\tau}^i(\mathbf{w}) = \frac{\text{exp}(\mathbf{w}^{T}\phi_i(\tau))} {\sum^{K-1}_{j=0}\text{exp}(\mathbf{w}^{T}\phi_j(\tau))} $ to represent the above expression compactly as

\begin{equation*}
    \big{|}  R(\widehat{\mathbf{w}}_n,\tau) -  R(\tau)\big{|} 
    \leq  K \sum_{i=0}^{K-1}\bigg{|} f_{\tau}^i(\mathbf{w}^*) - f_{\tau}^i(\widehat{\mathbf{w}}_n)\bigg{|}.
\end{equation*}
We can compute the gradient of $f_{\tau}^i(\mathbf{w})$ as follows:

$$\nabla f_{\tau}^i(\mathbf{w}) = \sum_{j=0}^{K-1} \frac{\text{exp}(\mathbf{w}^T\phi_j(\tau)) \; \text{exp}(\mathbf{w}^{T}\phi_i(\tau))}{\big{(}\sum_{l=0}^{K-1}\text{exp}(\mathbf{w}^T\phi_l(\tau)) \big{)}^2}\bigg(\phi_{i}(\tau)-\phi_j(\tau)\bigg)$$

$$\nabla f_{\tau}^i(\mathbf{w}) = \sum_{j=0}^{K-1} \frac{\text{exp}(\mathbf{w}^T(\phi_i(\tau) +\phi_j(\tau)) }{\big{(}\sum_{l=0}^{K-1}\text{exp}(\mathbf{w}^T\phi_l(\tau)) \big{)}^2}\bigg(\phi_{i}(\tau)-\phi_j(\tau)\bigg)$$
We  bound the norm of the gradient vector as follows:

$$||\nabla f_{\tau}^i(\mathbf{w})||_2 = \bigg{|}\bigg{|}\sum_{j=0}^{K-1} \frac{\text{exp}(\mathbf{w}^T(\phi_i(\tau) +\phi_j(\tau)) }{\big{(}\sum_{l=0}^{K-1}\text{exp}(\mathbf{w}^T\phi_l(\tau)) \big{)}^2}\bigg(\phi_{i}(\tau)-\phi_j(\tau)\bigg)\bigg{|}\bigg{|}_2$$

$$||\nabla f_{\tau}^i(\mathbf{w})||_2 \leq \sum_{j=0}^{K-1}\bigg{|}\bigg{|} \frac{\text{exp}(\mathbf{w}^T(\phi_i(\tau) +\phi_j(\tau)) }{\big{(}\sum_{l=0}^{K-1}\text{exp}(\mathbf{w}^T\phi_l(\tau)) \big{)}^2}\bigg(\phi_{i}(\tau)-\phi_j(\tau)\bigg)\bigg{|}\bigg{|}_2$$
According to assumption \ref{assumption::bounded_features}, we have 
$ \frac{w^T (\phi_i(\tau) + \phi_j(\tau))}{\left( \sum_{l=0}^{K-1} \exp(\mathbf{w}^T \phi_l(\tau)) \right)^2} \leq \frac{\exp(4B)}{K^2} $.

$$||\nabla f_{\tau}^i(\mathbf{w})  ||_2 \leq \frac{\text{exp(4B)}}{K^2} \sum_{j=0}^{K-1}\bigg{|}\bigg{|} \bigg(\phi_{i}(\tau)-\phi_j(\tau)\bigg)\bigg{|}\bigg{|}_2 \leq \frac{\text{exp(4B)}}{K^2} (2K) =  2\frac{\text{exp(4B)}}{K} $$
Thus, the function $f_{\tau}^i(\mathbf{w})$ is a differentiable map from $\mathbb{R}^{Kd}$ to $\mathbb{R}$, and the following equality holds from the fundamental theorem of calculus.

\begin{align*}
    f_{\tau}^i(\mathbf{w}^*) &= f_{\tau}^i(\widehat{\mathbf{w}}_n) + \int_{0}^{1} \nabla f_{\tau}^i(\widehat{\mathbf{w}}_n + t(\mathbf{w}^*-\widehat{\mathbf{w}}_n))^T(\mathbf{w}^*-\widehat{\mathbf{w}}_n) \, dt \\
 &= \int_{0}^{1} \nabla f_{\tau}^i(\widehat{\mathbf{w}}_n + t(\mathbf{w}^*-\widehat{\mathbf{w}}_n))^T(\mathbf{w}^*-\widehat{\mathbf{w}}_n) \, dt \\
    | f_{\tau}^i(\mathbf{w}^*) - f_{\tau}^i(\widehat{\mathbf{w}}_n) | &= \left| \int_{0}^{1} \nabla f_{\tau}^i(\widehat{\mathbf{w}}_n + t(\mathbf{w}^*-\widehat{\mathbf{w}}_n))^T(\mathbf{w}^*-\widehat{\mathbf{w}}_n) \, dt \right| \\
   &\leq \int_{0}^{1} \left| \nabla f_{\tau}^i(\widehat{\mathbf{w}}_n + t(\mathbf{w}^*-\widehat{\mathbf{w}}_n))^T(\mathbf{w}^*-\widehat{\mathbf{w}}_n) \right| \, dt \\
   &\leq \int_{0}^{1} \|\nabla f_{\tau}^i(\widehat{\mathbf{w}}_n + t(\mathbf{w}^*-\widehat{\mathbf{w}}_n))\|_2 \|\mathbf{w}^{*}-\widehat{\mathbf{w}}_n\|_2 \, dt \\
   &\leq \int_{0}^{1} 2\frac{\exp(4B)}{K} \|\mathbf{w}^{*}-\widehat{\mathbf{w}}_n\|_2 \, dt \\
&\leq 2\frac{\exp(4B)}{K} \|\mathbf{w}^{*}-\widehat{\mathbf{w}}_n\|_2
\end{align*}

The first inequality holds due to the triangular inequality, and the second one holds due to the Cauchy-Schwarz inequality. Now, recall that:

\begin{equation*}
    \big{|}  R(\widehat{\mathbf{w}}_n,\tau) -  R(\tau)\big{|} 
    \leq  K \sum_{i=0}^{K-1}\bigg{|} f_{\tau}^i(\mathbf{w}^*) - f_{\tau}^i(\widehat{\mathbf{w}}_n)\bigg{|}
\end{equation*}

\begin{equation*}
    \big{|}  R(\widehat{\mathbf{w}}_n,\tau) -  R(\tau)\big{|} 
    \leq  2K\text{exp(4B)} ||\mathbf{w}^{*}-\widehat{\mathbf{w}}_n||_2 
\end{equation*}
Using the high probability concentration result from Lemma \ref{lemm_concen}, we have the following for any possible trajectory $\tau \in \mathcal{T}$:

\begin{equation*}
    \big{|}  R(\widehat{\mathbf{w}}_n,\tau) -  R(\tau)\big{|} 
    \leq   
\frac{4K\text{exp(4B)}}{\eta \lambda_{\min}(\Sigma_{D_n})} \sqrt{\frac{C^2}{2n}\log{\frac{4}{\delta}}} \quad \text{w.p. at least}  \;\; 1-\delta
\end{equation*}

where  $\Sigma_{D_n} = \frac{1}{nK^2} \sum_{i=1}^n \sum_{j=0}^{K-1} \sum_{l=0}^{K-1}(\phi_j(\tau^{(i)})-\phi_l(\tau^{(i)}))(\phi_j(\tau^{(i)})-\phi_l(\tau^{(i)}))^T,
\eta = \frac{\exp(-4B)}{2} $ and $ C = \log{(K \exp(2B))}.$

\vspace{0.4em}

This completes the Proof of Lemma \ref{reward_concen}.

\subsection{Proof of Theorem \ref{thm1}}

The optimistic reward function is defined as follows: 

\begin{equation*}
    \overline{R}(\widehat{\mathbf{w}}_n,\tau) =  \min\bigg{(}R(\widehat{\mathbf{w}}_n,\tau) + \frac{4K\text{exp(4B)}}{\eta \lambda_{\min}(\Sigma_{D_n})} \sqrt{\frac{C^2}{2n}\log{\frac{4}{\delta}}},\;K-1\bigg{)}.
\end{equation*}
Recall that with initial state distribution $\rho$, the value function of a policy $V^{\pi}$ is defined as follows:

\begin{equation*}
    V^\pi := \mathbb{E}_{s_1 \sim \rho,\tau \sim \mathbb{P}^{\pi}(.|s_1)} [R(\tau)] = \mathbb{E}_{s_1 \sim \rho,\tau \sim \mathbb{P}^{\pi}(.|s_1)} [R(\tau)]
\end{equation*}
We construct the policy  at $n-$th episode $\pi^{(n)}$  as follows:

\begin{equation*}
    \pi^{(n)} \in \arg \max_{\pi \in \Pi} \;\mathbb{E}_{s_1 \sim \rho,\tau \sim \mathbb{P}^{\pi}(.|s_1)} [\overline{R}(\widehat{\mathbf{w}}_n,\tau)]
\end{equation*}
The optimistic value function for a policy $\pi$ at  episode $n$ is defined as follows:

\begin{equation*}
    \overline{V}^\pi_{n} := \mathbb{E}_{s_1 \sim \rho,\tau \sim \mathbb{P}^{\pi}(.|s_1)}  [\overline{R}(\widehat{\mathbf{w}}_n,\tau)]
\end{equation*}
The cumulative regret $R(N)$ is defined as follows:

\begin{equation*}
    R(N) := \sum_{i=1}^{N} V^{\pi^*} - V^{\pi^{(i)}}
\end{equation*}
Using the concept of optimism in the face of uncertainty, we can upper bound the value function of the optimal policy as follows:

\begin{align*}
   \sum_{i=1}^{N} V^{\pi^*} & = \sum_{i=1}^{N} \mathbb{E}_{s_1 \sim \rho,\tau \sim \mathbb{P}^{\pi^{*}}(.|s_1)} [R(\tau)] \\ & \leq \sum_{i=1}^{N}   \mathbb{E}_{s_1 \sim \rho,\tau \sim \mathbb{P}^{\pi^{*}}(.|s_1)} [\overline{R}(\widehat{\mathbf{w}}_i,\tau)] \quad 
   \text{w.p. at least } 1-N\delta  \\ & \leq \sum_{i=1}^{N}   \mathbb{E}_{s_1 \sim \rho,\tau \sim \mathbb{P}^{\pi^{(i)}}(.|s_1)} [\overline{R}(\widehat{\mathbf{w}}_i,\tau)]  \\ & = \sum_{i=1}^{N}   \overline{V}^{\pi^{(i)}}_{i}\
\end{align*}
The second inequality holds due to Lemma \ref{reward_concen} and the union bound, and the second inequality holds by the construction of the policy \( \pi^{(i)} \). Now, with probability at least \( 1 - N\delta \), we have the following:

\begin{equation}
    R(N) \leq \sum_{i=1}^{N} \overline{V}^{\pi^{(i)}}_{i} - V^{\pi^{(i)}} \quad   \text{w.p. at least } 1-N\delta
    \label{useful_res}
\end{equation}
We revisit the following lemma that bounds the sum of martingale difference sequence:

\begin{lemma} \cite{chatterji2021theory}
    Let $\left\{x_i\right\}_{i=1}^{\infty}$ be a martingale difference sequence with $\left|x_i\right| \leq \zeta$ and let $\delta \in(0,1]$. Then with probability $1-\delta$ for all $N \in \mathbb{N}$:
$$
\sum_{i=1}^N x_t \leq 2 \zeta \sqrt{N \log \left(\frac{6 \log N}{\delta}\right)}.
$$
\label{lemm_MDL}
\end{lemma}
Now, consider the following martingale difference sequence:

\begin{equation*}
 x_i :=  \overline{V}^{\pi^{(i)}}_{i} - V^{\pi^{(i)}} - \big{(}\; \overline{R}(\widehat{\mathbf{w}}_i,\tau^{(i)})- R(\tau^{(i)})\; \big{)}
\end{equation*}
Since both the reward function and the optimistic reward function are bounded by \( K-1 \), we have \( |x_i| \leq 2K \). Using Lemma \ref{lemm_MDL}, we then have the following with probability at least $1-\delta$:

\begin{equation*}
    \sum_{i=1}^{N} x_i = \sum_{i=1}^{N} \bigg {[} \; \overline{V}^{\pi^{(i)}}_{i} - V^{\pi^{(i)}} - \big{(}\; \overline{R}(\widehat{\mathbf{w}}_i,\tau^{(i)})- R(\tau^{(i)})\; \big{)} \bigg {]} \;\leq  4K \sqrt{N \log \left(\frac{6 \log N}{\delta}\right)}
\end{equation*}

\begin{equation*}
    \sum_{i=1}^{N} \overline{V}^{\pi^{(i)}}_{i} - V^{\pi^{(i)}}  \leq  \sum_{i=1}^{N} \big{(}\; \overline{R}(\widehat{\mathbf{w}}_i,\tau^{(i)})- R(\tau^{(i)})\; \big{)} +  4K \sqrt{N \log \left(\frac{6 \log N}{\delta}\right)}
\end{equation*}
We analyze the following sum:

\begin{align*}
      &\sum_{i=1}^{N} \big{(}\; \overline{R}(\widehat{\mathbf{w}}_i,\tau^{(i)})- R(\tau^{(i)}) \; \big{)}  
     \\ &  =  \sum_{i=1}^{N} \bigg{[}   \min\bigg{(}R(\widehat{\mathbf{w}}_i,\tau^{(i)}) + \frac{4K\text{exp(4B)}}{\eta \lambda_{\min}(\Sigma_{D_n})} \sqrt{\frac{C^2}{2i}\log{\frac{2}{\delta}}},\;K-1\bigg{)}  -\min(\;R(\tau^{(i)}),K-1\;)\bigg{]} \\
     &\leq   \sum_{i=1}^{N} \bigg{|} R(\widehat{\mathbf{w}}_i,\tau^{(i)}) + \frac{4K\text{exp(4B)}}{\eta \lambda_{\min}(\Sigma_{D_n})} \sqrt{\frac{C^2}{2i}\log{\frac{2}{\delta}}} - R(\tau^{(i)}) \bigg{|} \\
     &\leq \sum_{i=1}^{N} \bigg{|} R(\widehat{\mathbf{w}}_i,\tau^{(i)}) - R(\tau^{(i)}) \bigg{|} + \frac{4K\text{exp(4B)}}{\eta \lambda_{\min}(\Sigma_{D_n})} \sqrt{\frac{C^2}{2i}\log{\frac{2}{\delta}}} \\
      &\leq \sum_{i=1}^{N}  \frac{8K\text{exp(4B)}}{\eta \lambda_{\min}(\Sigma_{D_n})} \sqrt{\frac{C^2}{2i}\log{\frac{4}{\delta}}} \quad 
   \text{w.p. at least } 1-N\delta
\end{align*}
The first inequality holds due to the fact that the map \( x \mapsto \min(x, c) \) is 1-Lipschitz for any real constant \( c \). The second inequality holds due to the triangle inequality. The last inequality follows from Lemma \ref{reward_concen} with the application of the union bound. We define \( \lambda = \min_{i \in \{1,2,3,\dots,N\}} \lambda_{\min}(\Sigma_{D_n}) \). We have the following with probability at least $1-(N+1)\delta$.

\begin{equation*}
     \sum_{i=1}^{N} \overline{V}^{\pi^{(i)}}_{i} - V^{\pi^{(i)}} \leq \sum_{i=1}^{N}  \frac{8K\text{exp(4B)}}{\eta \lambda_{\min}(\Sigma_{D_n})} \sqrt{\frac{C^2}{2i}\log{\frac{4}{\delta}}} + 4K \sqrt{N \log \left(\frac{6 \log N}{\delta}\right)}
\end{equation*}

\begin{equation*}
     \sum_{i=1}^{N} \overline{V}^{\pi^{(i)}} - V^{\pi^{(i)}} \leq  \frac{8K\text{exp(4B)}}{\eta \lambda}  \sqrt{\frac{C^2}{4}\log{\frac{2}{\delta}}} \sum_{i=1}^{N}  \sqrt{\frac{1}{i}} + 4K \sqrt{N \log \left(\frac{6 \log N}{\delta}\right)}
\end{equation*}
We bound the sum $\sum_{i=1}^{N}  \sqrt{\frac{1}{i}}$ as follows:

\begin{equation*}
    \sum_{i=1}^{N}  \sqrt{\frac{1}{i}} \leq \int_{0}^{N}  \sqrt{\frac{1}{x}} dx = 2\sqrt{N}
\end{equation*}
Now, with probability at least \( 1 - (N+1)\delta \), we have the following:

\begin{align*}
        \sum_{i=1}^{N} \overline{V}^{\pi^{(i)}} - V^{\pi^{(i)}} & \leq  \frac{8K\text{exp(4B)}}{\eta \lambda}  \sqrt{\frac{C^2}{2}\log{\frac{2}{\delta}}} \; \bigg{(} 2\sqrt{N}\bigg{)} + 4K \sqrt{N \log \left(\frac{6 \log N}{\delta}\right)}
        \\ & = \frac{16K\text{exp(4B)}}{\eta \lambda}  \sqrt{\frac{NC^2}{2}\log{\frac{4}{\delta}}}  + 4K \sqrt{N \log \left(\frac{6 \log N}{\delta}\right)}
\end{align*}
Alternatively, since \( N \geq 1 \), we can say that with probability at least \( 1 - 2N\delta \), we have the following:

\begin{align*}
        \sum_{i=1}^{N} \overline{V}^{\pi^{(i)}} - V^{\pi^{(i)}} \leq \frac{16K\text{exp(4B)}}{\eta \lambda}  \sqrt{\frac{NC^2}{2}\log{\frac{4}{\delta}}}  + 4K \sqrt{N \log \left(\frac{6 \log N}{\delta}\right)}.
\end{align*}
Using the result from Equation \ref{useful_res} and the union bound, we now have the high-probability bound on the cumulative regret, i.e., the following holds with probability at least \( 1 - 3N\delta \):

 \begin{align*}
       R(N) \leq  \sum_{i=1}^{N} \overline{V}^{\pi^{(i)}} - V^{\pi^{(i)}} \leq \frac{16K\text{exp(4B)}}{\eta \lambda}  \sqrt{\frac{NC^2}{2}\log{\frac{4}{\delta}}}  + 4K \sqrt{N \log \left(\frac{6 \log N}{\delta}\right)}.
\end{align*}
If we set \( \delta' \) = \( 3N\delta \), it follows that the following holds with probability at least \( 1 - \delta' \):

\begin{align*}
       R(N) \leq  \frac{16K\text{exp(4B)}}{\eta \lambda}  \sqrt{\frac{NC^2}{2}\log{\frac{12N}{\delta'}}}  + 4K \sqrt{N \log \left(\frac{18N \log N}{\delta'}\right)}  
\end{align*}
where 
\[
\eta = \frac{\exp(-4B)}{2}, \quad \text{and} \quad C = \log{(K \exp(2B))}.
\]

This completes the Proof of Theorem \ref{thm1}.

\newpage
\subsection{Practical Implementation of the K-UCBVI}

\begin{algorithm}
\hrule
\vspace{0.3em}
\caption{K-UCBVI with Multilevel Episodic Feedback and Known Transition Probabilities (Practical Implementation)}
\vspace{0.20em}
\hrule
\vspace{0.3em}
Initialize $\theta$, $\epsilon$, $\widehat{\mathbf{w}}$\\
\For{$n = 1, ..., N$}{
\While{$\|\theta_{\text{new}} - \theta_{\text{old}} \| \geq \epsilon$}{
\begin{equation*}
    \pi_{\theta}(a|s) = \Bigg[\frac{\text{exp} \; \theta_{a,s}}{\sum_{a'\in A} \text{exp} \; \theta_{a',s}}\Bigg]\\
\end{equation*}

\begin{equation*}
    \nabla \overline{V}^{\pi_{\theta}} = \mathbb{E}_{\tau \sim P_{\rho}^{\pi_{\theta}}}\Bigg[\widehat{R}_{t}(\tau) \sum_{t=0}^H \nabla \text{log} (\pi_{\theta}(a|s) \Bigg]\\
\end{equation*}

\begin{equation*}
    \theta_{\text{new}} = \theta_{\text{old}} + \alpha \nabla \overline {V}^{\pi_{\theta}}
\end{equation*}

Sample a trajectory from the current optimal policy $\pi^{ \theta_{\text{new}}}(a|s)$
\\
Get a new feedback score $y$ for the sampled trajectory. \\
Solve for $\widehat{\mathbf{w}}$

}
}

\vspace{0.2em}
\hrule
\vspace{0.3em}
\label{algo:2}
\end{algorithm}



\end{document}